\documentclass[letterpaper]{article} 
\usepackage{aaai23}  
\usepackage{times}  
\usepackage{helvet}  
\usepackage{courier}  
\usepackage[hyphens]{url}  
\usepackage{graphicx} 
\usepackage{natbib}  
\usepackage{caption} 
\usepackage{algorithm}
\usepackage{algorithmic}
\usepackage{multirow}
\usepackage{bigstrut}
\usepackage{booktabs}
\usepackage{array}
\newcommand{\etal}{{\emph{et al. }}}

\usepackage{amsmath}
\usepackage{amssymb}
\usepackage{balance}
\usepackage{bm} 
\usepackage{color}
\usepackage[colorlinks=true,citecolor=blue]{hyperref}

\usepackage{newfloat}
\usepackage{listings}
\DeclareCaptionStyle{ruled}{labelfont=normalfont,labelsep=colon,strut=off} 
\floatstyle{ruled}
\newfloat{listing}{tb}{lst}{}
\floatname{listing}{Listing}
\nocopyright 

\title{Spatial-Spectral Transformer for Hyperspectral Image Denoising}
\author {
Miaoyu Li \textsuperscript{\rm 1},
Ying Fu \textsuperscript{\rm 1} \thanks{Corresponding author},
Yulun Zhang \textsuperscript{\rm 2}
}
\affiliations {
\textsuperscript{\rm 1} Beijing Institute of Technology,  
\textsuperscript{\rm 2} ETH Z\"{u}rich\\
\{miaoyli, fuying\}@bit.edu.cn, yulun100@gmail.com
}

\usepackage{bibentry}

\begin{document}

\maketitle
\begin{abstract}
	
	Hyperspectral image (HSI) denoising is a crucial preprocessing procedure for the subsequent HSI applications. Unfortunately, though witnessing the development of deep learning in HSI denoising area, existing convolution-based methods face the trade-off between computational efficiency and capability to model non-local characteristics of HSI. In this paper, we propose a Spatial-Spectral Transformer (SST) to alleviate this problem. To fully explore intrinsic similarity characteristics in both spatial dimension and spectral dimension, we conduct non-local spatial self-attention and global spectral self-attention with Transformer architecture. The window-based spatial self-attention focuses on the spatial similarity beyond the neighboring region. While, spectral self-attention exploits the long-range dependencies between highly correlative bands. Experimental results show that our proposed method outperforms the state-of-the-art HSI denoising methods in quantitative quality and visual results. The code is released at \textcolor{magenta}{https://github.com/MyuLi/SST}.

\end{abstract}

\section{Introduction}

Hyperspectral images (HSIs) provide abundant information in spectral dimension  
and have been widely applied to the fields of remote sensing~\cite{cloutis1996review}, material recognition~\cite{thai2002invariant}, agriculture~\cite{kersting2012pre}, medical diagnosis~\cite{fei2020hyperspectral} and so on.  
However, in the sensing process, due to limited light, photon effects, and atmospheric interference, HSIs often suffer from corruption and noise,  which negatively influences the subsequent HSI applications. Therefore, HSI denoising is a critical pre-processing procedure to enhance image quality for the aforementioned high-level computer vision tasks. 

Compared to color images, HSIs offer pixel-level spectral features by imaging narrow spectral bands over a continuous spectral range. It means there are statistical similarities between bands. Early denoising works, such as dictionary learning method~\cite{elad2006image}, 
and BM3D \cite{dabov2007image}, focused on the non-local similarity in spatial dimension but did not take the spectral features into account. Thus, more HSI denoising works exploit both the spatial similarity and spectral correlation.
Multilinear tools were used in~\cite{renard2008denoising} to extract spectral components and spatial
information for denoising. The parallel factor analysis model was employed in~\cite{liu2012denoising} to exploit the decomposition uniqueness
and single rank character of HSIs. Low-rank prior~\cite{chen2017denoising,chang2020weighted}, sparse representation~\cite{zhuang2018fast}, and  total variation regularization~\cite{du2018joint} have also been widely adopted to HSI denoising. Despite the various hand-crafted priors,  traditional model-based HSI denoising methods are always hard to optimize and time-consuming.

With the development of deep learning, convolutional neural networks (CNNs) based HSI denoising methods~\cite{yuan2018hyperspectral,chang2018hsi,dong2019deep,zhang2019hybrid,shi2021hyperspectral} have shown distinguished advantages over traditional HSI denoising methods. CNN-based methods rely on convolution filters to model the data dependencies in spatial dimension and spectral dimension, facing the trade-off between computational efficiency and the ability to model non-local similarity of HSIs. Besides, the learned convolution filters are with static weights, which means the filters used for feature extraction are fixed in the testing phase. The denoising process is based on the knowledge learned from training dataset, but, the inner characteristics of the target HSI are not fully exploited. 

Recently, Transformer models have been applied to vision tasks~\cite{zhang2020feature,carion2020end,dai2021up}. Transformers apply the self-attention mechanism~\cite{wang2018non} across image regions and can well capture the internal similarity of target image. According to previous analysis~\cite{cai2022mask,zamir2022restormer}, Transformer could be a powerful alternative to CNNs in HSI denoising task. However, existing Transformer-based works mainly aim at natural images with ignorance in spectral similarity. Since HSIs have strong spectral correlations, such neglect would negatively affect denoising results. A practical way of employing Transformer to HSI denoising is applying spectral-wise attention
to well utilize spectral correlation.

In this work, we propose a Spatial-Spectral Transformer to sufficiently explore the non-local spatial similarity and global spectral correlation of HSIs. Firstly, the spatial information of HSI is restored by the shifted window-based self-attention, which conducts a non-local spatial-wise coarse denoising beyond neighboring pixels. Secondly, the high correlation between bands is exploited by spectral-wise self-attention, which conducts globally weighted denoising on each pixel with fine details. Finally, the output from self-attention module is passed through the multi-layer perceptron (MLP) and skip-connection for smooth convergence.
In summary, the main contributions of our work are as follows:
\vspace{-2.5mm}
\begin{itemize}
	\setlength{\parsep}{0pt}
	\setlength{\parskip}{0pt}
	\item We propose a Spatial-Spectral Transformer to fully exploit both the  non-local spatial similarity and global spectral correlation of target noisy HSI.
	\vspace{1mm}
	\item  We design an efficient denoising module by integrating window-based spatial self-attention with spectral self-attention. The coarse spatial features are finely weighted by spectral attention to obtain detailed features. 
	\vspace{1mm}
	\item Extensive experimental results on various noise degradations show that our proposed method outperforms state-of-the-art methods in terms of both objective metrics and subjective visual quality.
\end{itemize}

\section{Related Work}
In this section, we briefly review two major research directions which are related to our work, including the latest progress in HSI denoising and vision Transformers.

\subsection{HSI Denoising}
Existing HSI denoising methods could be roughly classified into two categories, including traditional model-based methods and deep learning-based methods. 

Model-based HSI denoising methods usually utilize hand-crafted prior knowledge of HSIs. The non-local similarity \cite{zhang2019hybrid}, total variation \cite{he2015total,yuan2012hyperspectral}, low-rank \cite{zhang2013hyperspectral,chang2017hyper,he2021tslrln,chang2020weighted}, and sparse representation \cite{lu2015spectral,zhuang2018fast} are frequently used to exploit the intrinsic spatial and spectral characteristics of HSIs. Fu \etal \cite{fu2015adaptive} presented an adaptive spatial-spectral dictionary learning method 
and high correlations in both domains are exploited. In \cite{he2019non}, 
the spatial non-local similarity and global spectral low-rank
property were integrated for HSI denoising. Zhang \etal~\cite{zhang2021double} proposed a double low-rank matrix decomposition method. These methods generally formulated HSI denoising as a complex iterative problem and required a long time to optimize.

Existing learning-based HSI denoising methods~\cite{chang2018hsi,yuan2018hyperspectral,zhang2019hybrid,sidorov2019deep} rely on deep CNNs to automatically learn the prior from large-scale datasets. In \cite{chang2018hsi}, a spatial-spectral deep residual CNN was employed for HSI restoration. To exploit the global dependency and correlation information in both dimensions, Shi \etal \cite{shi2021hyperspectral} proposed dual-attention denoising network that could obtain more essential feature extraction of HSI. Wei \etal \cite{wei20203} designed a 3D quasi-recurrent neural network
(QRNN3D) to make full use of structural spatial-spectral
correlation as well as global correlation along the spectrum. In \cite{bodrito2021trainable}, a hybrid trainable spectral-spatial sparse coding model was proposed.

Although these CNN-based HSI denoising methods have
achieved excellent performance, data similarities in spectral dimension is always underestimated. Moreover, with trained parameters fixed, convolution filters follow a stereotyped paradigm to extract features and lack the adaptability to exploit the intrinsic similarity characteristic of noisy HSI.

\subsection{Vision Transformers}

Transformer was firstly proposed in \cite{vaswani2017attention} for NLP tasks and then was successfully applied to numerous vision tasks, such as image classification~\cite{zhang2020feature}, image segmentation~\cite{wang2021end}, object detection~\cite{dai2021up}, and face recognition~\cite{zhong2021face}. Generally, Transformer contains three essential components, including LayerNorm (LN) module, self-attention module, and multi-layer perceptron (MLP) module. When Transformer is first applied to visual tasks, the whole image was treated as a sequence of non-overlapping
medium-sized image patches in ViT \cite{dosovitskiy2020image}. To  make Transformer more efficient, pyramid vision Transformer (PVT) was proposed in~ \cite{wang2021pyramid}. 
Swin Transformer~ \cite{liu2021swin} utilized a shift operation and had linear computational complexity to image size. In \cite{zamir2022restormer}, an efficient Transformer for high-resolution image restoration was proposed.

Though Transformer has achieved excellent performance, existing Transformers mainly focus on exploiting spatial similarity and lack exploration of spectral correlation. Thus, directly applying previous Transformers to the HSI task \cite{zhong2021spectral,pang2022trq3dnet} may be less effective in exploiting two-dimensional features of HSIs.

\section{Method}
\begin{figure*}[t]
	\scriptsize
	\centering	
	\includegraphics[width=1\linewidth]{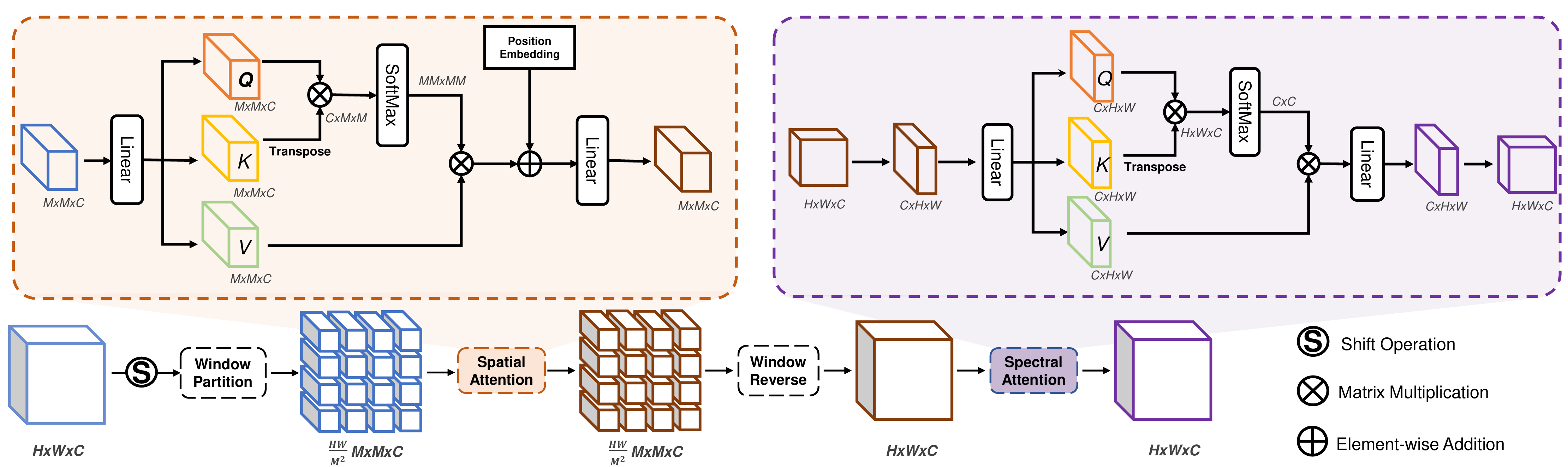}
	\vspace{-5mm}                         
	\caption{Illustration of Spatial-Spectral Multi-head self-Attention. The SSMA module mainly contains non-local spatial multi-head self-attention and global spectral multi-head self-attention.}\label{fig:SS-MTA}
\end{figure*}

In this section, we first illustrate the problem formulation of HSI denoising and the motivation of our work. Then, we describe the proposed spatial-spectral multi-head self-attention module in detail (see Figure \ref{fig:SS-MTA}). Finally, the
overall architecture of our proposed SST is provided (see Figure \ref{fig:overall}).

\subsection{Motivation and Formulation}
Mathematically, the additive noise degradation model for one HSI could be formulated as follows
\begin{equation}
	\label{noise_model}
	\bm{Y}=\bm{X}+ \bm{\eta},
\end{equation}
where $\bm{X}\in\mathbb{R}^{H\times W\times B}$ stands for the clean HSI, and $\bm{Y}\in\mathbb{R}^{H\times W\times B}$ is the noisy HSI corrupted by additive noise $\bm{\eta}$. $H$ and $W$ stand for spatial height and width, respectively. $B$ denotes the spectral resolution of HSI. 

HSIs in the real-world are usually degraded by multifarious noise, including Gaussian noise, stripe noise, impulse noise, deadline noise, or a mixture of them \cite{zhang2013hyperspectral,chang2020weighted}. The goal of HSI denoising is to restore the desired clean HSI $\bm{X}$ from noisy observation $\bm{Y}$.

The effectiveness of non-local spatial similarity prior has been verified by previous traditional model-based HSI denoising works~\cite{zhang2019hybrid,chang2020weighted}. Since these methods use hand-crafted priors, they lack the exploration of statistical information from external datasets. Fortunately, the spatial self-attention mechanism could provide a comparative ability to obtain the non-local similarity of images. What is more, with learnable parameters to obtain diverse feature expression, 
a network integrated with self-attention operation can automatically learn the deep prior knowledge from large-scale datasets, and obtains better HSI denoising results. Thus, it is natural to apply the spatial self-attention operation to learning-based HSI denoising.

Moreover, HSI is usually regarded as a data cube and is especially rich in spectral information. The similarities exist not only in the spatial dimension but also in the spectral dimension. Since most existing vision Transformers only conduct spatial self-attention to get representative spatial features, the comprehensive consideration of spatial dimension and spectral dimension is lacking.
Therefore, in this work, we combine spectral self-attention with spatial self-attention to exploit both the non-local spatial similarity and spectral high correlation of HSIs. With shifted local spatial window attention, the non-local neighborhood information beyond the pixel is well exploited with computation efficiency. With global spectral attention, the spectral correlation is well modeled and utilized to benefit the denoising process.

\subsection{Spatial-Spectral Transformer}
In this section, we first introduce our proposed spatial-spectral Transformer layer (SSTL) with our designed self-attention module. To effectively exploit the non-local spatial similarity and spectral correlation of target HSI, we propose a spatial-spectral multi-head self-attention (SSMA) module for HSI denoising. Moreover, according to \cite{dong2021attention}, skip connections and MLP are beneficial to prevent the network from falling into rank collapse. Thus, we follow the structure proposed in \cite{dosovitskiy2020image} with self-attention module followed by MLP layer, residual connections, and LayerNorm.

Concretely, let $\bm{Z}_{l-1}\in \mathbb{R}^{H\times W\times C}$ stands for the input feature embeddings of the $l$-th SSTL, the overall processing structure of SSTL could be expressed as: 
\begin{equation}
	\label{eq:sstl}
	\begin{aligned}
		&\bm{Z}_{l}^{\prime} = {\rm SSMA}({\rm LN}(\bm{Z}_{l-1})) + \bm{Z}_{l-1},\\
		&\bm{Z}_{l} = {\rm MLP}({\rm LN}(\bm{Z}_{l}^{\prime})) + \bm{Z}_{l}^{\prime},
	\end{aligned}
\end{equation}
where $\bm{Z}_{l}^{\prime}$ and $\bm{Z}_{l}$ denote the outputs of SSMA and SSTL.

The details of our SSMA are illustrated in Figure \ref{fig:SS-MTA}. It mainly includes a non-local spatial self-attention (NLSA) layer and a global spectral self-attention (GSA) layer. 

Given normalized input features, $ \bm{Z}^{in}\in \mathbb{R}^{H\times W\times C}$, a window partition operation is first conducted in spatial dimension with a window size of $M$. Thus, the whole input features are divided into $\frac{HW}{M^{2}}$ non-overlapping patches as $\left\lbrace \bm{Z}_{1}^{in}, ..., \bm{Z}_{i}^{in}, ... ,\bm{Z}_{I}^{in}\right\rbrace$, where $ \bm{Z}_{i}^{in} \in \mathbb{R}^{{M^2}\times C}$. After partition, each patch $ \bm{Z}_{i}^{in} $ individually passes through the NLSA layer to exploit the non-local similarity in the spatial dimension. This process can be expressed as follows:
\begin{align}
	&\left\lbrace \bm{Z}^{in}_i\right\rbrace  = {\rm WinPartition}\left(\bm{Z}^{in} \right), i=1,...,I
	\label{eq:local_i}\\& \bm{Z}^{ns}_{i} = {\rm NLSA}\left( \bm{Z}^{in}_{i}\right), i=1,...,I\\
	&\bm{Z}^{ns} = {\rm WinReverse}\left( \left\lbrace  \bm{Z}^{ns}_{i} \right\rbrace \right), i=1,...,I.
\end{align}

After gathering output patches from NLSA layer together through the window reverse operation, the obtained features $\bm{Z}^{ns} $ are fed directly to GSA layer. The global spectral correlation is well utilized and helps the removal of noise as:
\begin{align}
	\label{eq:global}&\bm{Z}^{gs} = { \rm GSA}\left(  \bm{Z}^{ns}\right) .
\end{align}

\noindent\textbf{Non-Local Spatial self-Attention (NLSA).}
For real-world HSI denoising 
applications, there is a fundamental trade-off between model capability
and model flexibility. Though global spatial self-attention could bring spatial long-range dependencies in a large perspective field, it is frustratingly not suitable for the HSI denoising task since its quadratic complexity grows with spatial size. Thus, we employ self-attention with shifted window to obtain the internal non-local spatial similarity information beyond neighbors. Consequently, NLSA layer provides considerable ability to express spatial features in linear complexity.

For NLSA layer, each input patch $\bm{Z}^{in}_{i} \in \mathbb{R}^{ M^2\times C}$ is linearly projected into \textit{query} ${\bm{Q}_{i}^{ns}}$, \textit{key} $ {\bm{K}_{i}^{ns}}$, and \textit{value} $\bm{V}_{i}^{ns}\in{\mathbb{R}^{ M^2\times C}}$ to increase the representation ability as:
\begin{equation}
	\bm{Q}_{i}^{ns} = \bm{Z}^{in}_{i}\bm{W}_{q}^{ns}, 	\bm{K}^{ns} =  \bm{Z}^{in}_{i}\bm{W}_{k}^{ns}, 	\bm{V}^{ns} =  \bm{Z}^{in}_{i}\bm{W}_{v}^{ns}, 
\end{equation}
where $\bm{W}_{q}^{ns},\bm{W}_{k}^{ns}$, and $\bm{W}_{v}^{ns}$ are weights of size $C\times C$.

Subsequently, to jointly assemble information from different representation subspaces, multi-head mechanism is employed. $\bm{Q}_{i}^{ns}, \bm{K}_{i}^{ns}$, and $\bm{V}_{i}^{ns}$ are split into $N$ heads as $\bm{Q}_{ij}^{ns}$, $\bm{K}_{ij}^{ns}$, and $ \bm{V}_{ij}^{ns}$, respectively. Thus, the non-local spatial self-attention matrix for each $ head_{ij}^{ns}$ is computed as:
\begin{equation}
	\begin{aligned}
		&\bm{A}^{ns}_{ij} = {\rm Softmax}(\bm{Q}_{ij}^{ns}{\bm{K}^{ns}_{ij}}^{T}/\sqrt{d}+\bm{B}),\\
		&head^{ns}_{ij} = \bm{A}^{ns}_{ij}\bm{V}_{ij}^{ns},
	\end{aligned}
\end{equation}
where $d$ is the dimension of $\bm{Q}_{ij}^{ns}$, specifically as $C/N$. And $B$ is the relative bias defined in \cite{liu2021swin}. After self-attention operation, the outputs $head^{ns}_{ij}$ are embedded together and linearly projected to get $\bm{Z}_{i}^{ns}$ in Eq. \eqref{eq:local_i}. Moreover, we also conduct a spatial shift operation between each SSTL to obtain more comprehensive spatial information. It can bring interactions between local windows. Specifically, the shift operation is conducted by shifting the input features by $\lfloor |\frac{M}{2}|,|\frac{M}{2}|\rfloor $ pixels before partitioning.

\begin{figure}[t]
	\scriptsize 
	\centering	
	\includegraphics[width=1.0\linewidth]{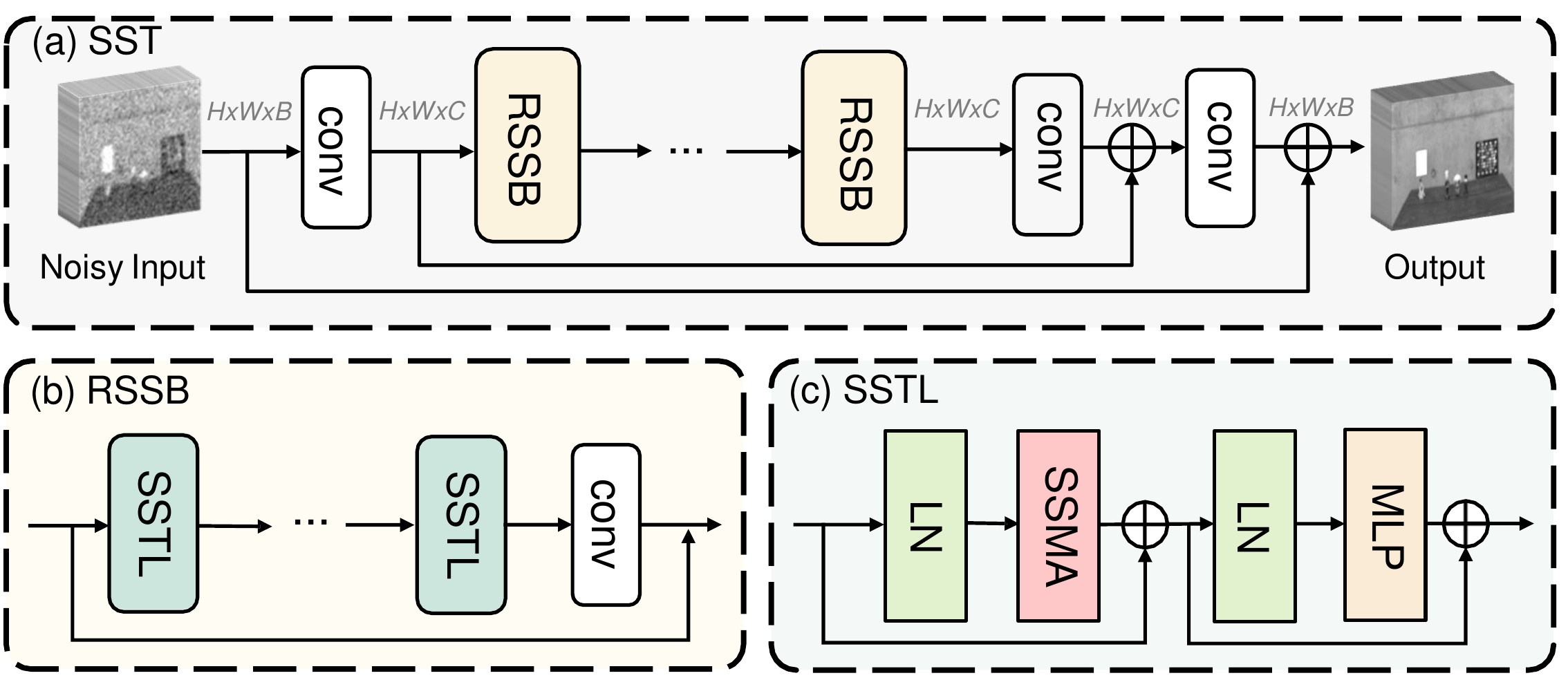}
	\vspace{-4.2mm}                                  
	\caption{Overall architecture of SST. (a) The basic pipeline of SST. (b) Residual spatial-spectral block (RSSB). (c) Spatial-spectral Transformer layer (SSTL).}
	\label{fig:overall}
	\vspace{-2mm}
\end{figure}

\noindent\textbf{Global Spectral self-Attention (GSA).}
After the non-local spatial self-attention operation, the features are already spatially representative for HSI. But it still lacks spectral representations. Since HSI has underlying spectral correlations, it provides sufficient similarity information for the self-attention operation to obtain long-range dependencies. Thus, we employ global spectral self-attention after non-local spatial self-attention. By doing so, the non-local spatial similarity and spectral correlation are both finely considered. 

For a single GSA layer, given an input from NSLA layer, $\bm{Z}^{ns}$$\in$$\mathbb{R}^{H\times W\times C}$, the input is firstly transposed and reshaped into $\bm{Z}^{T}$$\in$$\mathbb{R}^{C\times HW} $. Then, it is also linearly projected to $\bm{Q}\in{\mathbb{R}^{C\times HW}}$, $\bm{K}\in{\mathbb{R}^{ C\times HW}}$, and $\bm{V}^{gs}\in{\mathbb{R}^{ C\times HW}}$ as:
\begin{equation}
	\bm{Q}^{gs} = \bm{W}_{q}^{gs}\bm{Z}^{T}, 	\bm{K}^{gs} = \bm{W}_{k}^{gs}\bm{Z}^{T},  	\bm{V}^{gs} = \bm{W}_{v}^{gs}\bm{Z}^{T}, 
\end{equation}
where $\bm{W}_{q},\bm{W}_{k},\bm{W}_{v}$ are weights of size $C\times C$.

Similar to non-local spatial attention, we split $\bm{Q}^{gs},\bm{K}^{gs}$, and $\bm{V}^{gs}$ into $N$ heads. Each $head^{gs}_{j}$  is defined by:
\begin{equation}
	\begin{aligned}
		&\bm{A}^{gs}_{j} = {\rm Softmax}({\bm{K}^{gs}_{j}}^{T}{\bm{Q}_{j}}^{gs}/\sqrt{d}),\\
		&head^{gs}_{j} = \bm{V}_{j}^{gs}\bm{A}^{gs}_{j}.
	\end{aligned}
\end{equation}

The outputs $head^{gs}_{j}$ are concatenated in spectral dimension and projected to $\bm{Z}^{gs}$ in Eq. \eqref{eq:global}. It is worth emphasizing that $\bm{Z}^{gs}$ contains more spectral details without losing critical spatial information compared to $\bm{Z}^{ns}$. 

\noindent\textbf{Computational Complexity.}
We analyze the computational complexity of our proposed SSMA module as:
\begin{equation}
	\begin{aligned}
		&\mathcal{O}\left( NLSA\right)=\left({M^2}HWC \right), \mathcal{O}\left( GSA\right)=\left(C^{2}HW\right), \\
		&\mathcal{O}\left(SSMA\right)=\left({M^2}HWC+C^{2}HW\right) , 
	\end{aligned}
\end{equation}
where $M$ and $C$ are predefined constants. Thus,  our proposed SSMA module achieves linear computation cost.

\noindent\textbf{Overall Network Architecture.} As shown in Figure \ref{fig:overall}, our whole Transformer first employs one 3$\times$3 convolution layer to extract low-level features embeddings $\bm{F}_0 \in \mathbb{R}^{H\times W\times C}$ from noisy observation $\bm{Y}$$\in$$\mathbb{R}^{H\times W\times B}$. Then, shallow features pass through $T$ Residual Spatial-Spectral Block (RSSB) layers with a fixed feature size of $H$$\times$$W$$\times$$C$. The process of deep feature extraction could be denoted as:
\begin{equation}
	\label{denoise}
	\bm{F}_t = H_t(\bm{F}_{t-1}), t = 1,2,...,T,
\end{equation}
where ${H}_t$ is the $t$-th RSSB layer and $F_t$ stands for the $t$-th spatial-spectral feature map obtained by RSSB layer.

The designed RSSB contains 6 Spatial-Spectral Transformer layers.
Each of RSSB ends up with a 3$\times$3 convolution layer. We have adopted a skip connection that adds  residual features from the previous block.

To recover from deep feature $\bm{F}_T$, two 3$\times$3 convolution layers are concatenated
with shallow features via skip connections. With a global skip connection that adds the noisy input to the output, the middle network actually learns a noise pattern that matches the noise distribution of input.

\section{Experiments}
In this section, we first introduce the datasets and settings used in our experiments. The quantitative metrics and competing methods are also covered. Then, we provide denoising results quantitatively and qualitatively on simulated data and real data. Finally, ablation studies are carried out to analyze the effectiveness of the  components in our model.

\subsection{Simulated Experiments}
\noindent\textbf{Datasets.}
We evaluate our method mainly on ICVL~\cite{arad2016sparse} dataset. It consists of 201 images with 1392×1300 spatial resolution and 31 spectral bands from 400 nm to 700 nm. We follow the dataset settings in \cite{bodrito2021trainable}, which uses 100 HSIs for training and 50 HSIs for testing to ensure pictures captured from the same scene are only used once. Specifically, we center
crop training images to size 1024$\times$1024 and normalize them to $[0, 1]$. Then, we extract patches of size 64$\times$64 at different scales, with strides 64, 32, and 32. As for testing samples, each HSI is cropped to size 512$\times$512$\times$31 for better visual effects. Normalization is also conducted on testing HSIs.

\begin{table*}[t]
	\centering
	\resizebox{\textwidth}{!}{
		\setlength{\tabcolsep}{0.8mm}
		\begin{tabular}{l|c|c|c|c|c|c|c|c|c|c|c|c|c|c|c}
			\hline
			\multirow{2}[4]{*}{Method} & \multicolumn{3}{c|}{10} & \multicolumn{3}{c|}{30} & \multicolumn{3}{c|}{50} & \multicolumn{3}{c|}{70} & \multicolumn{3}{c}{10-70} \bigstrut\\
			\cline{2-16}        & PSNR & SSIM  & SAM   & PSNR  & SSIM  & SAM  & PSNR   & SSIM  & SAM  & PSNR  & SSIM   & SAM  & PSNR & SSIM  & SAM \bigstrut\\
			\hline
			\hline
			Noisy & 28.13 & 0.879 & 18.72 & 18.59 & 0.552 & 37.9  & 14.15 & 0.348 & 49.01 & 11.23 & 0.230 & 56.45 & 17.24 & 0.478 & 41.94 \bigstrut\\
			\hline
			BM4D~\cite{maggioni2012nonlocal}& 40.78 & 0.993 & 2.99  & 37.69 & 0.987 & 5.02  & 34.96 & 0.985 & 6.81  & 33.15 & 0.955 & 8.40  & 36.62 & 0.977 & 5.51 \bigstrut\\
			\hline
			LLRT~\cite{chang2017hyper}& 46.72 & 0.998 & 1.60  & 41.12 & 0.992 & 2.52  & 38.24 & 0.983 & 3.47  & 36.23 & 0.973 & 4.46  & 40.06 & 0.986 & 3.24 \bigstrut\\
			\hline
			TSLRLN\cite{he2021tslrln}& 46.07 & 0.998 & 1.82  & 41.26 & 0.994 & 3.03  & 38.37 & 0.989 & 4.36  & 36.44 & 0.983 & 5.69  & 40.22 & 0.991 & 3.58 \bigstrut\\
			\hline
			NGMeet~\cite{he2019non}& 47.90 & 0.999 & 1.39  & 42.44 & 0.982 & 2.06  & 39.69 & 0.966 & 2.49  & 38.05 & 0.953 & 2.83  & 41.67 & 0.994 & 2.19 \bigstrut\\
			\hline
			HSID-CNN \cite{yuan2018hyperspectral}& 43.14 & 0.992 & 2.12  & 40.30 & 0.985 & 3.14  & 37.72 & 0.975 & 4.27  & 34.95 & 0.952 & 5.84  & 39.04 & 0.978 & 3.71 \bigstrut\\
			\hline
			QRNN3D \cite{wei20203} &45.61 & 0.998 & 1.80  & 42.18 & 0.996 & 2.21  & 39.70 & 0.9924 & 3.00  & 38.09 & 0.988 & 3.42  & 41.34 & 0.994 & 2.42 \bigstrut\\
			\hline
			T3SC \cite{bodrito2021trainable} & 45.81 & 0.998 & 2.02  & 42.44 & 0.996 & 2.44  & 40.39 & 0.993 & 2.85  & 38.80 & 0.990 & 3.26  & 41.64 & 0.994 & 2.61 \bigstrut\\
			\hline
			SST (Ours)  & \textbf{48.28} & \textbf{0.999} & \textbf{1.30} & \textbf{43.32} & \textbf{0.997} & \textbf{1.87} & \textbf{41.09} & \textbf{0.995} & \textbf{2.19} & \textbf{39.55} & \textbf{0.992} & \textbf{2.46} & \textbf{42.57} & \textbf{0.996} & \textbf{1.99} \bigstrut\\
			\hline
		\end{tabular}%
	}
	\vspace{-2mm}
	\caption{Denoising comparisons under Gaussian noise with known variance on ICVL dataset. The best results are in bold.}
	\label{tab:guassian_fixed}%
	\vspace{-1mm}
\end{table*}%

\begin{figure*}
	\scriptsize
	\centering
	\setlength{\tabcolsep}{0.05cm}
	\begin{minipage}[t]{0.13\linewidth}
		\vspace{0mm}
		\begin{tabular}{c}
			\includegraphics[width=2.05\linewidth]{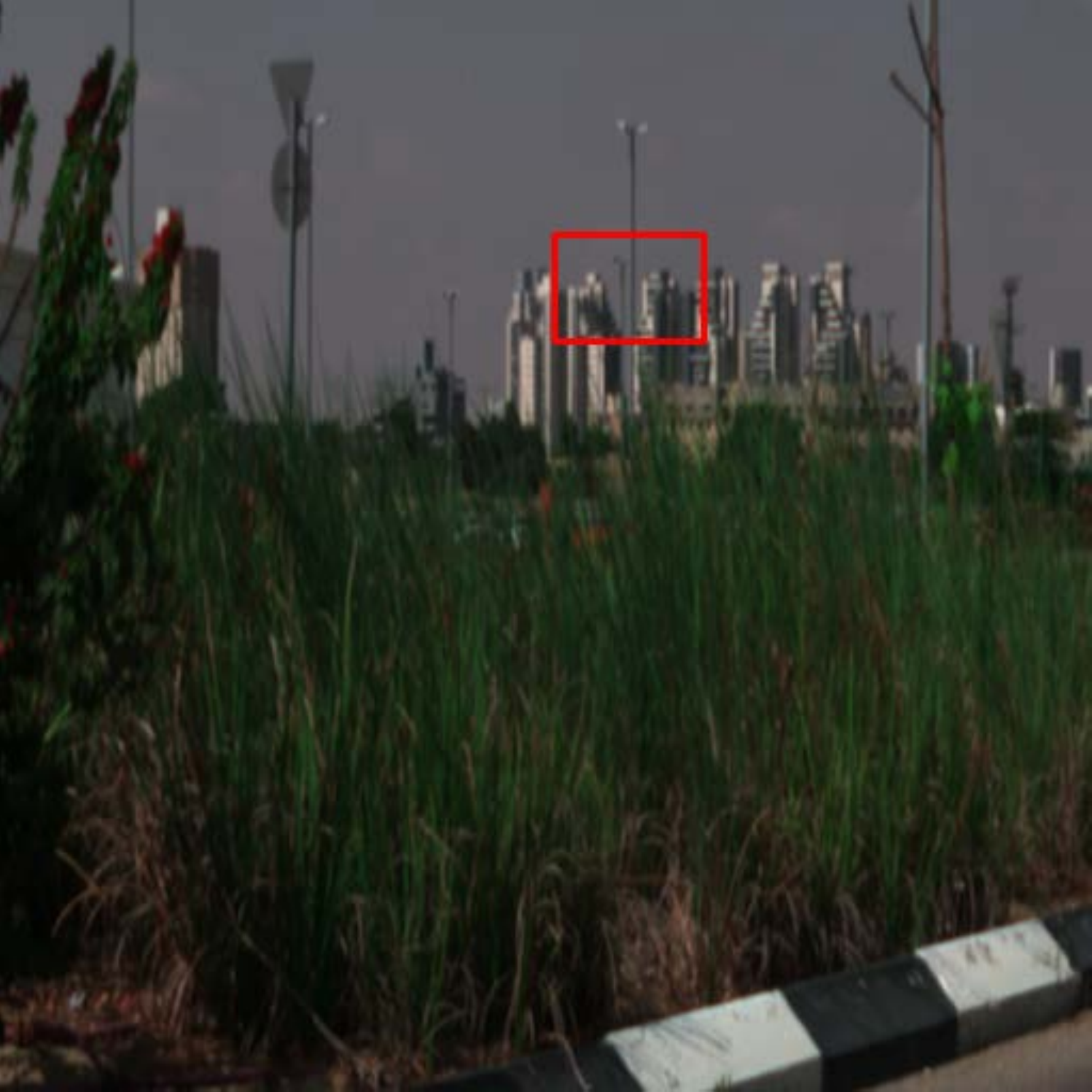}\\
			Gavyam\_0823-0950-1 from ICVL
		\end{tabular}
	\end{minipage}
	\begin{minipage}[t]{0.8\linewidth}
		\flushright
		\vspace{0mm}
		\begin{tabular}{ccccccc}
			\includegraphics[width=1.12in]{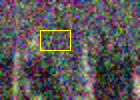}
			&\includegraphics[width=1.12in]{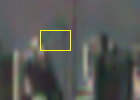}
			&\includegraphics[width=1.12in]{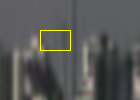}
			&\includegraphics[width=1.12in]{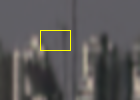}
			\\
			Noisy &  BM4D&LLRT  &NGMeet\\
			(14.15/0.4417) &  (33.42/0.9786)&(36.27/0.9832) &(37.95/0.9938)\\
			
			\includegraphics[width=1.12in]{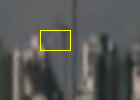}
			&\includegraphics[width=1.12in]{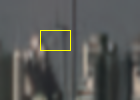}
			&\includegraphics[width=1.12in]{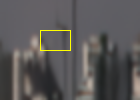}
			&\includegraphics[width=1.12in]{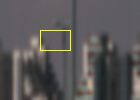}
			\\
			QRNN3D &T3SC&
			SST (Ours) & GroundTruth
			\\  (38.08/0.9940)&(38.39/0.9942) &(39.29/0.9955)&(PSNR/SSIM)\\
		\end{tabular}
	\end{minipage}
	\vspace{-2mm}
	\caption{Visual quality comparison under Gaussian noise level $\sigma$$=$$50$ on ICVL dataset using pseudo color image.}
	\label{fig:de_re} 
\end{figure*}

\noindent\textbf{Benchmark Models.} 
We compare our proposed method with four traditional methods, including BM4D \cite{maggioni2012nonlocal}, LLRT \cite{chang2017hyper}, TSLRLN \cite{he2021tslrln}, and NG-Meet \cite{he2019non}. Three deep learning methods are also used for comparison, including QRNN3D \cite{wei20203}, HSID-CNN \cite{yuan2018hyperspectral}, and T3SC \cite{bodrito2021trainable}. It is worth mentioning that HSID-CNN method traversed the noisy HSI through a one-by-one manner, specifically, inputs of the network are the current noisy band and its adjacent bands.

\noindent\textbf{Metrics.}
To quantitatively evaluate our proposed method, we employ three performance metrics, including peak signal-to-noise ratio (PSNR), structure
similarity (SSIM) \cite{wang2004image}, and spectral angle mapper
(SAM) \cite{yuhas1993determination}.  Larger values of PSNR and
SSIM imply better performance, while smaller values of SAM indicate the high fidelity of denoising results.

\noindent\textbf{Noise Patterns.} Following same settings in previous HSI denoising works~\cite{wei20203,bodrito2021trainable}, we evaluate our method on different noise patterns: 
\begin{itemize}
	\setlength{\itemsep}{0pt}
	\setlength{\parsep}{0pt}
	\setlength{\parskip}{0pt}
	\item i.i.d Gaussian noise with known variance $\sigma$ from 10 to 70. The noise level is the same on all bands.
	\item Unknown non-i.i.d Gaussian noise with variance $\sigma$ from  10 to 70. Different bands contain different level of noise.
	\item Non-i.i.d Gaussian noise with deadline noise, impulse noise, stripe noise, or mixture of them. The detailed settings could be found in \cite{wei20203}.
\end{itemize}

\begin{table*}[t]
	\centering
	\resizebox{\textwidth}{!}{
		\setlength{\tabcolsep}{0.8mm}
		\begin{tabular}{l|c|c|c|c|c|c|c|c|c|c|c|c|c|c|c}
			\hline
			\multirow{2}[4]{*}{Method} & \multicolumn{3}{c|}{Non-i.i.d} & \multicolumn{3}{c|}{Deadline} & \multicolumn{3}{c|}{Impulse} & \multicolumn{3}{c|}{Stripe} & \multicolumn{3}{c}{Mixture} \bigstrut\\
			\cline{2-16}          & PSNR  & SSIM & SAM   & PSNR & SSIM  & SAM  & PSNR  & SSIM  & SAM  & PSNR  & SSIM  & SAM & PSNR  & SSIM   & SAM\bigstrut\\
			\hline
			\hline
			Noisy & 18.29 & 0.512 & 46.20 & 17.50 & 0.477 & 47.55 & 14.93 & 0.376 & 46.98 & 17.51 & 0.487 & 46.98 & 13.91 & 0.340 & 51.53 \bigstrut\\
			\hline
			BM4D~\cite{maggioni2012nonlocal}& 36.18 & 0.977 & 5.78  & 33.77 & 0.962 & 6.85  & 29.79 & 0.861 & 21.59 & 35.63 & 0.973 & 6.26  & 28.01 & 0.842 & 23.59 \bigstrut\\
			\hline
			LLRT~\cite{chang2017hyper} & 34.18 & 0.962 & 4.88  & 32.98 & 0.956 & 5.29  & 28.85 & 0.882 & 18.17 & 34.27 & 0.963 & 4.93  & 28.06 & 0.870 & 19.37 \bigstrut\\
			\hline
			TSLRLN~\cite{he2021tslrln}& 41.95 & 0.994 & 2.89  & 36.56 & 0.977 & 5.52  & 33.72 & 0.893 & 22.89 & 38.41 & 0.985 & 4.99  & 27.25 & 0.852 & 25.23 \bigstrut\\
			\hline
			NGMeet~\cite{he2019non}& 34.90 & 0.975 & 5.37  & 33.41 & 0.967 & 6.55  & 27.02 & 0.788 & 31.20 & 34.88 & 0.967 & 5.42  & 26.13 & 0.778 & 31.89 \bigstrut\\
			\hline
			HSID-CNN~\cite{yuan2018hyperspectral} & 39.28 & 0.982 & 3.80  & 38.33 & 0.978 & 3.99  & 36.21 & 0.966 & 5.48  & 38.09 & 0.977 & 4.59  & 35.30 & 0.959 & 6.29 \bigstrut\\
			\hline
			QRNN3D~\cite{wei20203}& 42.18 & 0.995 & 2.84  & 41.69 & 0.994 & 2.61  & 40.32 & 0.991 & 4.31  & 41.68 & 0.994 & 2.97  & \textbf{39.08} & 0.989 & 4.80 \bigstrut\\
			\hline
			T3SC~\cite{bodrito2021trainable}  & 41.52 & 0.994 & 3.10  & 39.01 & 0.992 & 5.16  & 36.96 & 0.980 & 7.71  & 40.92 & 0.993 & 3.42  & 34.68 & 0.973 & 8.92 \bigstrut\\
			\hline
			SST (Ours)  & \textbf{43.57} & \textbf{0.997} & \textbf{2.05} & \textbf{42.74} & \textbf{0.996} & \textbf{2.21} & \textbf{41.66} & \textbf{0.995} & \textbf{2.60} & \textbf{42.97} & \textbf{0.996} & \textbf{2.15} & 38.78 & \textbf{0.991} & \textbf{2.99} \bigstrut\\
			\hline
	\end{tabular}}%
	\vspace{-2mm}
	\caption{Denoising comparisons under five complex noise cases on ICVL dataset. The best results are in bold.}
	\label{tab:comlex_noise}%
	\vspace{-1mm}
\end{table*}%

\begin{figure*}
	\scriptsize
	\centering
	\setlength{\tabcolsep}{0.05cm}
	\renewcommand{\arraystretch}{0.8}
	\begin{minipage}[t]{0.13\linewidth}
		\vspace{0mm}
		\begin{tabular}{c}
			\includegraphics[width=2.05\linewidth]{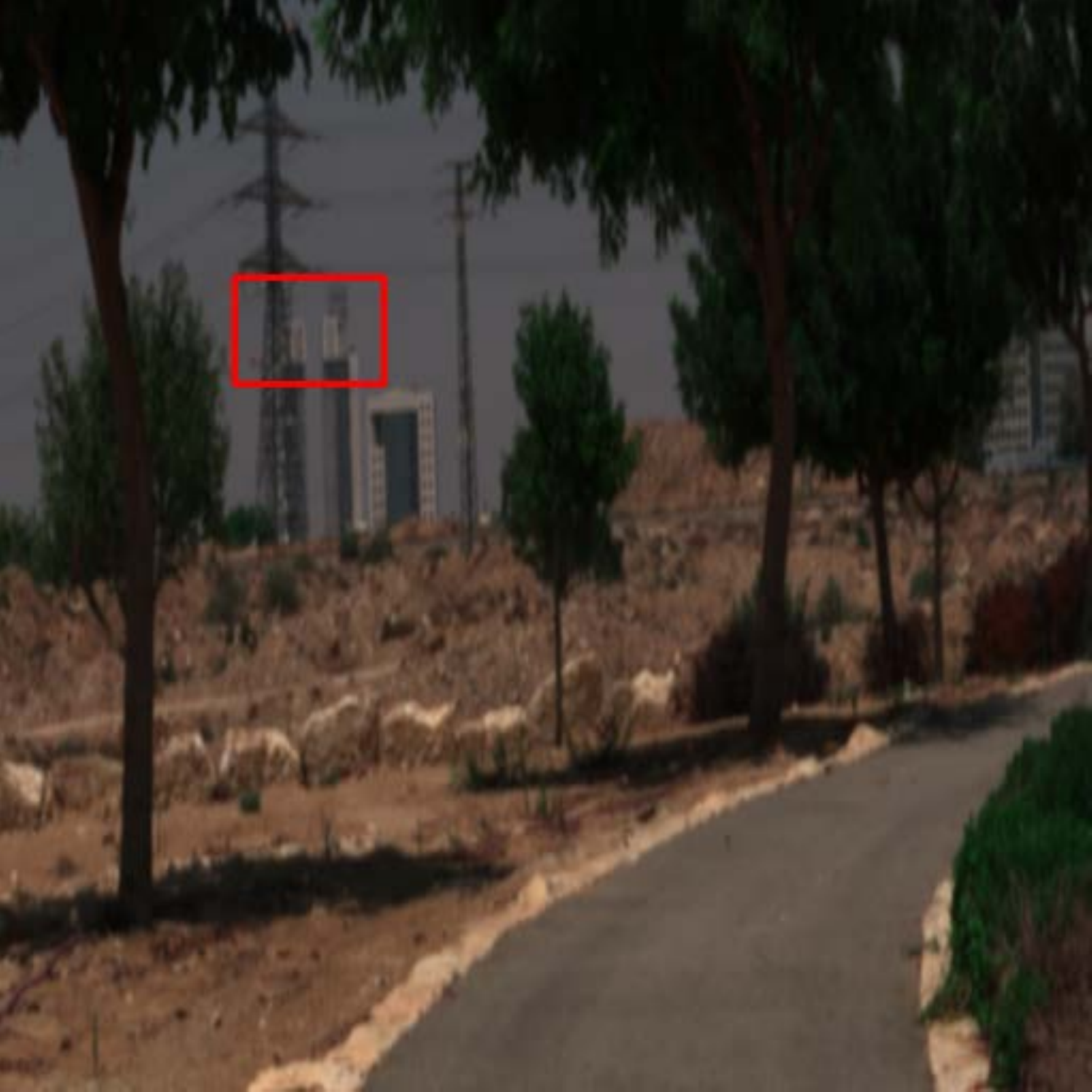}\\
			Nachal\_0823-1152 from ICVL
		\end{tabular}
	\end{minipage}
	\begin{minipage}[t]{0.8\linewidth}
		\flushright
		\vspace{0mm}
		\begin{tabular}{ccccccc}
			\includegraphics[width=1.12in]{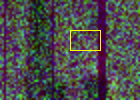}
			&\includegraphics[width=1.12in]{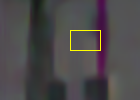}
			&\includegraphics[width=1.12in]{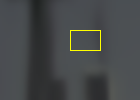}
			&\includegraphics[width=1.12in]{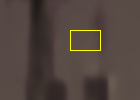}
			\\
			Noisy &  BM4D&LLRT  &NGMeet\\
			(18.26/0.5703) &  (33.06/0.9706)&(31.91/0.9581) &(32.09/0.9681)\\
			
			\includegraphics[width=1.12in]{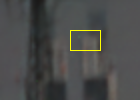}
			&\includegraphics[width=1.12in]{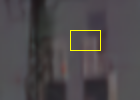}
			&\includegraphics[width=1.12in]{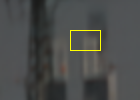}
			&\includegraphics[width=1.12in]{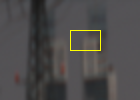}
			\\
			QRNN3D &T3SC&
			SST (Ours) & GroundTruth
			\\ 
			(38.08/0.9940) &  (38.39/0.9942)&(39.29/0.9955) &(PSNR/SSIM)\\
		\end{tabular}
	\end{minipage}
	\vspace{-2mm}
	\caption{Visual quality comparison under deadline noise on ICVL dataset using pseudo color image.}
	\label{fig:de_complex} 
\end{figure*}

\noindent\textbf{Implementation Details.}
We use Adam~\cite{kingma2014adam} to optimize the network with parameters initialized by Xavier initialization~\cite{glorot2010understanding}. The batch size is set to 8 with 100 epochs of training. The learning rate is set to 1$\times${10}$^{-4}$ and is divided by $10$ after 60 epoch. 
Competing deep learning methods (HSID-CNN, QRNN3D, and T3SC) and our proposed Transformer are implemented with PyTorch and run with a GeForce RTX 3090. 
Traditional methods, including BM4D, LLRT, TSLRLN, and NG-Meet, are implemented with Matlab and run with  an Intel Core i9-10850K CPU. All parameters involved in these competing algorithms were optimally assigned or automatically chosen as described in the reference papers.

\noindent\textbf{Gaussian Noise with Known Variance.}
In this case, zero mean additive white Gaussian noises with different variance $\sigma$ are added to the HSI to generate the noisy observations. 

The quantitative results of our method and compared methods on ICVL dataset are shown in Table \ref{tab:guassian_fixed}. Our method significantly outperforms all compared methods. With a noise level of $\sigma$$=$$70$, our method increases the PSNR by more than 0.7 dB. Furthermore, it can be seen that compared deep learning methods can achieve comparable results to traditional methods under high noise levels, but they are less effective to handle HSIs with low noise. Our proposed Transformer still achieve the best results under low level noise, showing its robustness and generalization ability. 

To demonstrate the denoising performance of our method, we show one denoised HSI from
different methods under $\sigma$$=$$50$ in Figure \ref{fig:de_re}. 
To further illustrate the result of the spectral fidelity, we use pseudo color images that is composed of bands 9, 15 and 28 for the red, green, and blue channels. 
BM4D method results in spatial discontinuity and loses many high frequency patterns. NG-Meet and LLRT obtain relatively good results, but they still lose fine textures as shown in the detailed figures. The results of QRNN3D and T3SC have obvious artifacts near the edge of the building. Our method obtains the most satisfying restored image along the spatial dimension and spectral dimension.

\noindent\textbf{Complex Noise with Unknown Non-i.i.d Gaussian Noise.}
Five types of complex
noise are added to generate noisy samples, including non-i.i.d Gaussian noise,  non-i.i.d Gaussian + deadline, non-i.i.d
Gaussian + impulse, non-i.i.d Gaussian + stripe, and mixture of them.
Quantitative results of our method and compared methods on ICVL dataset are shown in Table~\ref{tab:comlex_noise}. the visual comparison under non-i.i.d Gaussian noise + deadline noise case are shown in Figure~\ref{fig:de_complex}. It could be concluded that our method outperforms other methods under various types of noise degradation. An interesting observation is that the traditional methods almost all fail to restore a clean HSI under impulse noise and mixed noise, while the deep learning methods achieve considerable results. We speculate that the introduction of impulse noise and mixed noise to HSI makes it lose certain informative characteristics across the spatial dimension and spectral dimension.  Without the guidance of clean HSI, the hand-crafted prior is not strong enough to work well on noisy HSI. Our method not only focuses on the similarity information of HSI, but also has better adaptability to extract features through training, thus obtaining better results.
\begin{figure}
	\small
	\centering
	\setlength{\tabcolsep}{0.05cm}
	\begin{tabular}{ccccc}
		\includegraphics[width=0.78in]{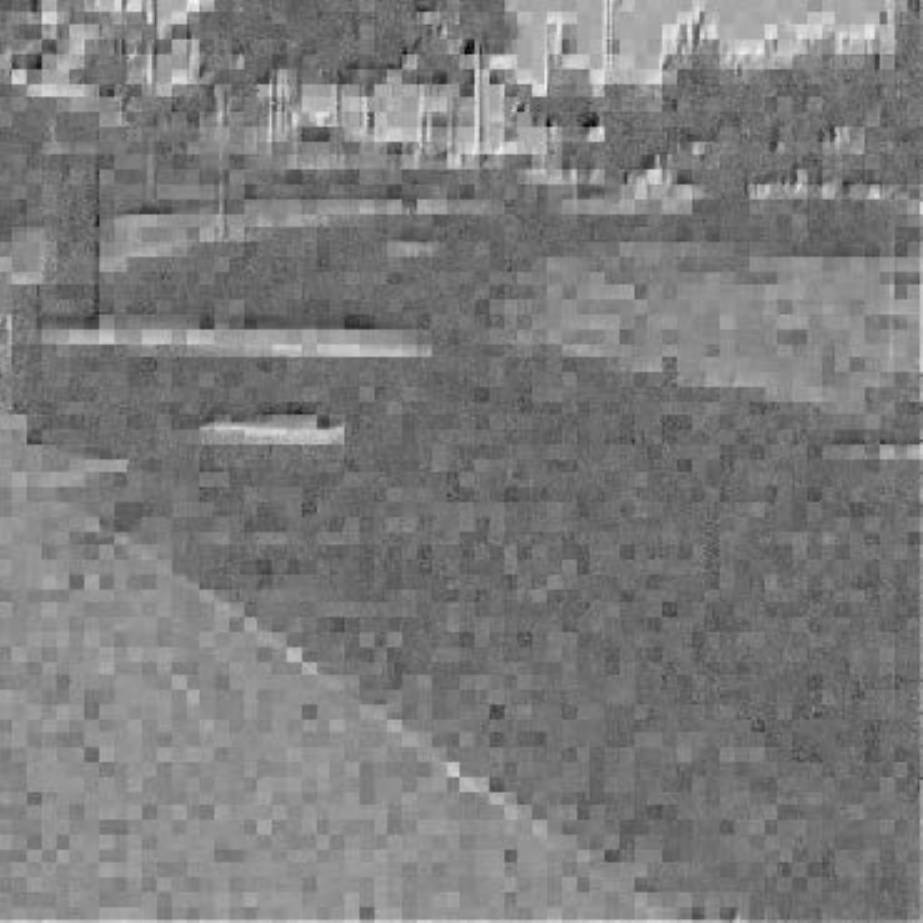}
		& \includegraphics[width=0.78in]{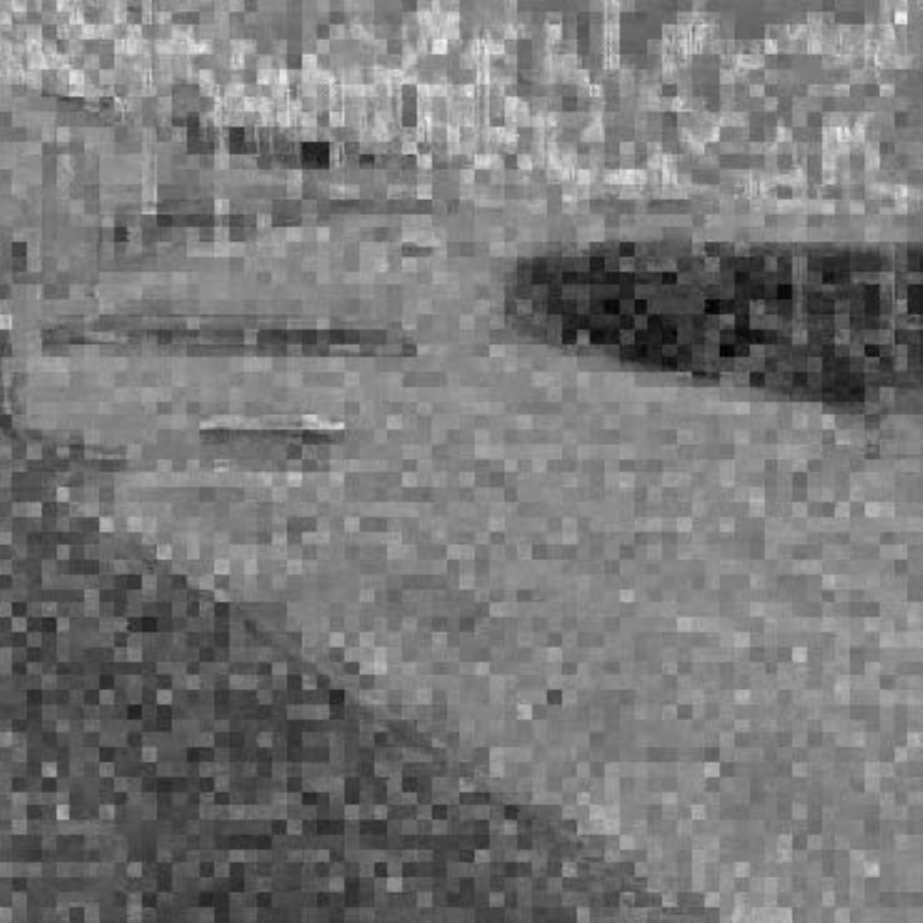}
		& \includegraphics[width=0.78in]{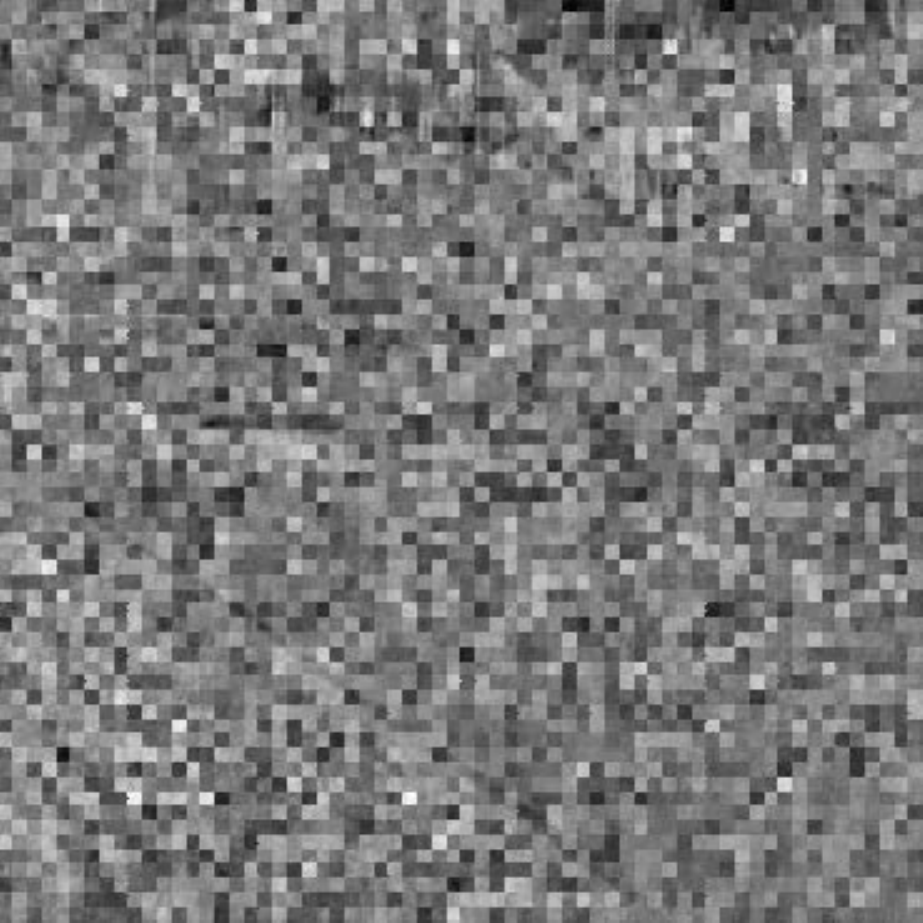}
		& \includegraphics[width=0.78in]{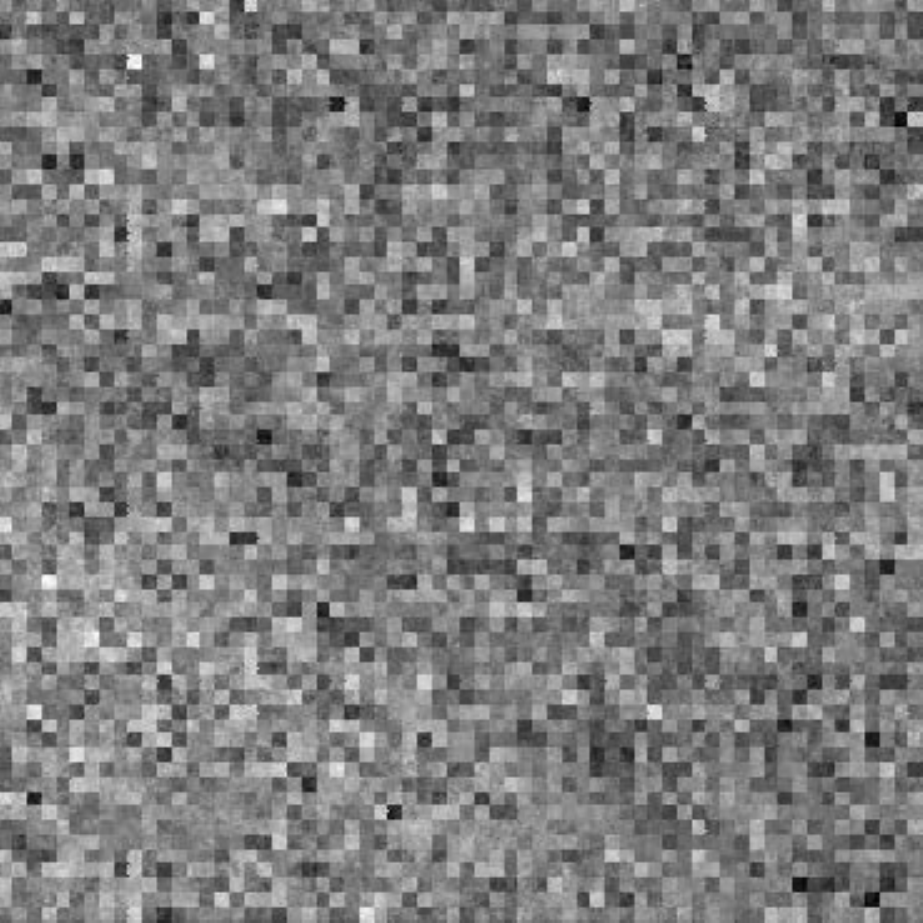}\\
		\multicolumn{4}{c}{(a) Outputs of NLSA at layer 1, 3, 5, and 6}
		\\
		\includegraphics[width=0.78in]{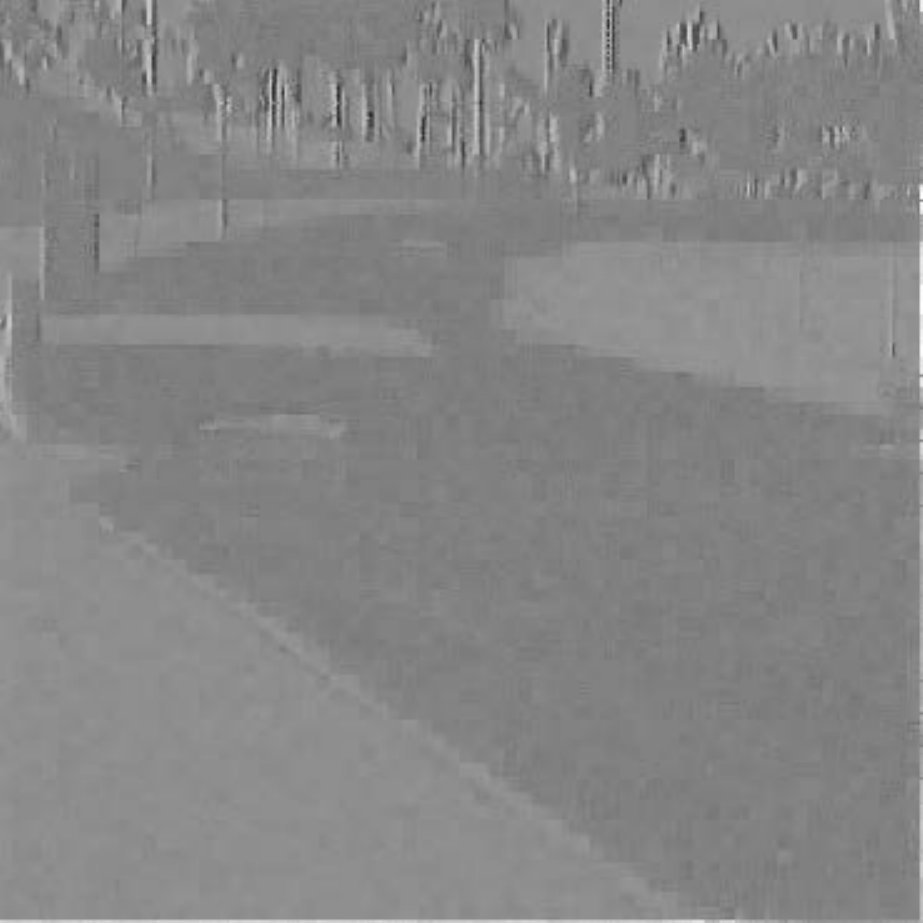}
		& \includegraphics[width=0.78in]{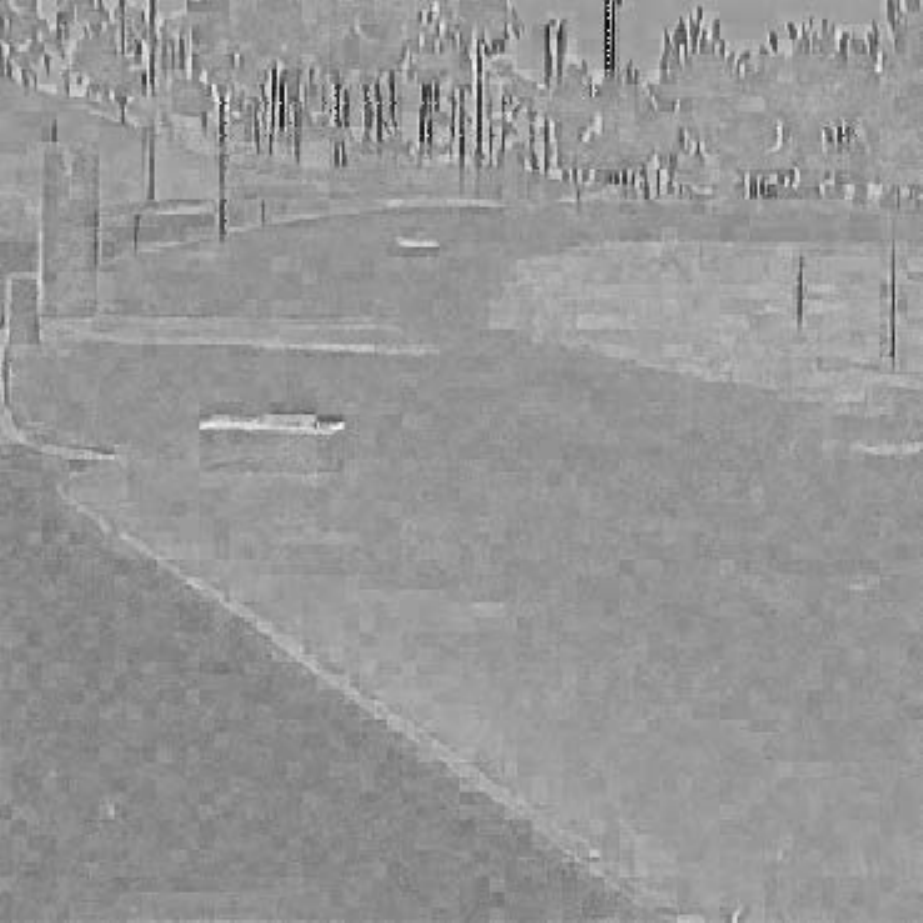}
		& \includegraphics[width=0.78in]{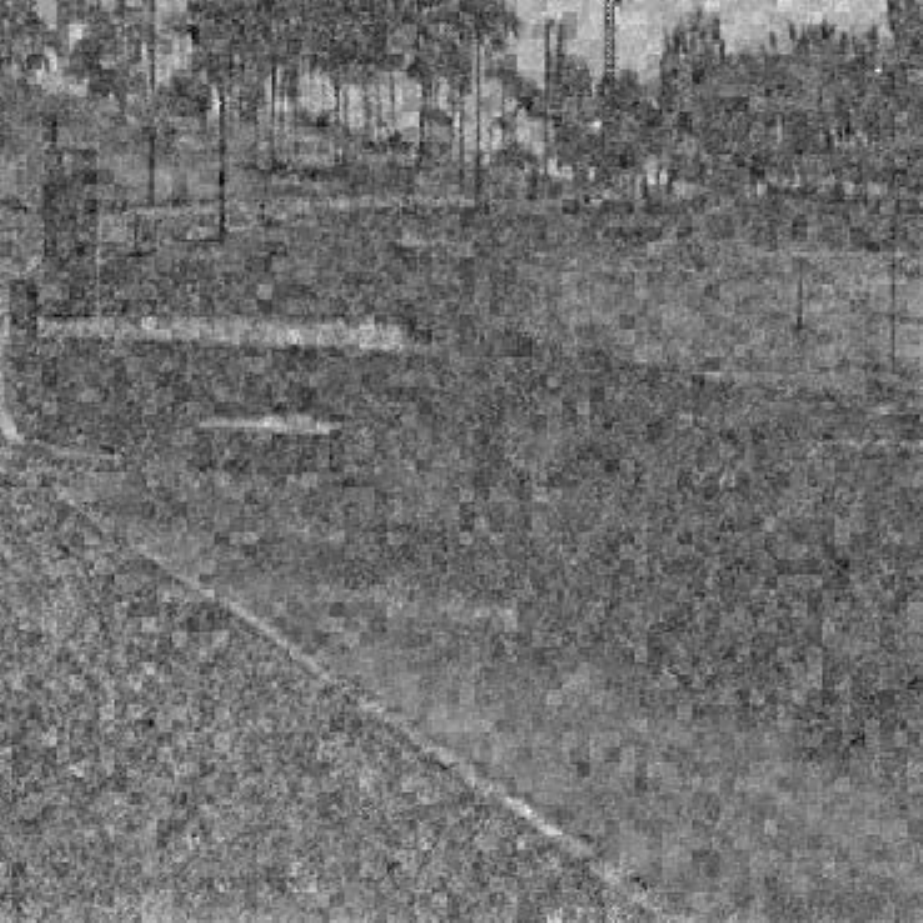}
		& \includegraphics[width=0.78in]{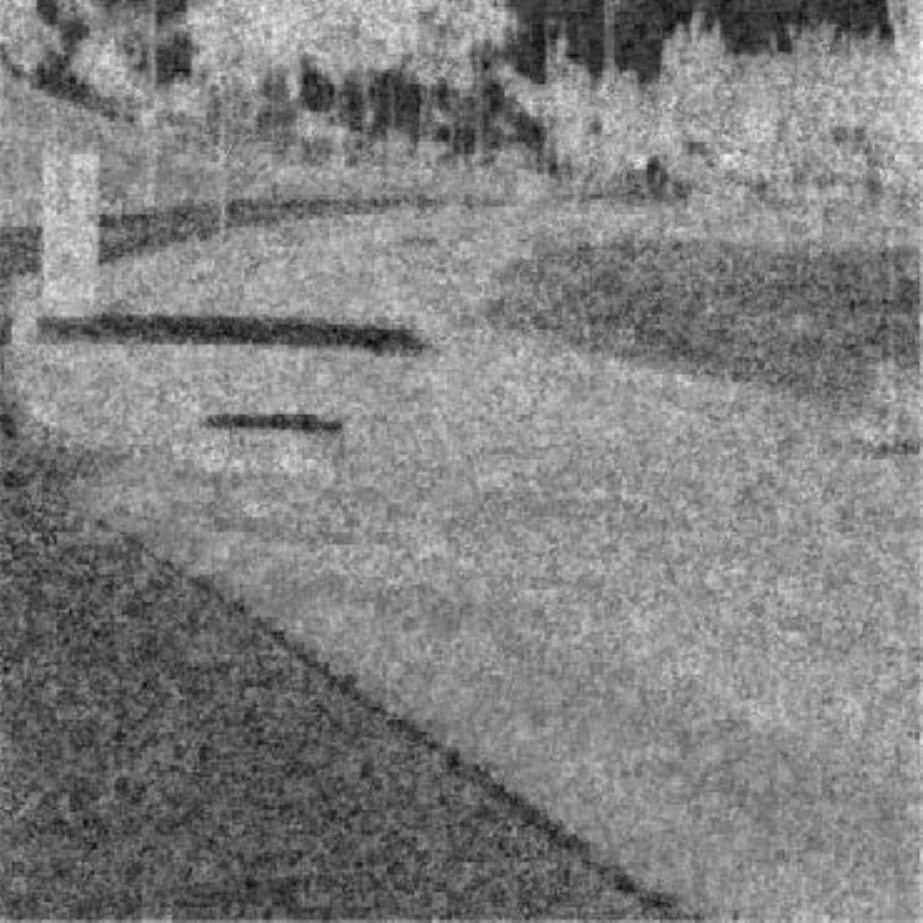}
		\\
		
		\multicolumn{4}{c}{(b) Outputs of GSA at layer 1, 3, 5, and 6} \\
	\end{tabular}
	\vspace{-2.5mm}
	\caption{Feature maps before and after GSA module. }
	\label{fig:feature}
\end{figure}\begin{figure*}
	\scriptsize
	\centering
	\setlength{\tabcolsep}{0.05cm}
	\begin{tabular}{ccccccc}
		\includegraphics[width=0.92in]{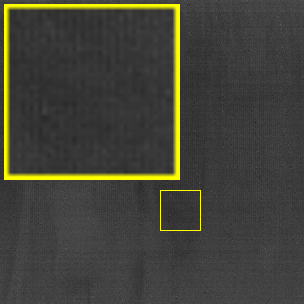}
		&\includegraphics[width=0.92in]{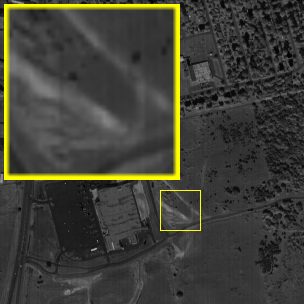}
		&\includegraphics[width=0.92in]{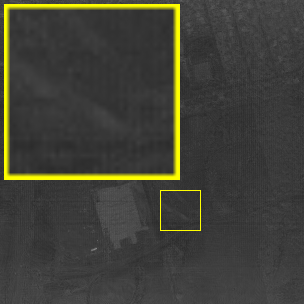}
		&\includegraphics[width=0.92in]{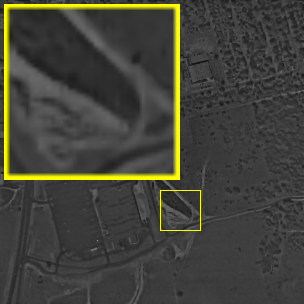}
		&\includegraphics[width=0.92in]{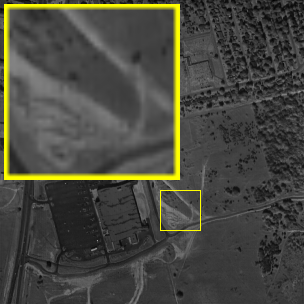}
		&\includegraphics[width=0.92in]{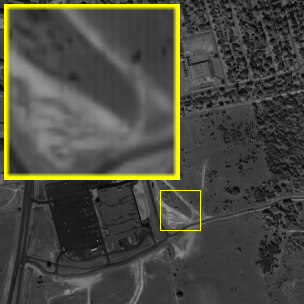}
		&\includegraphics[width=0.92in]{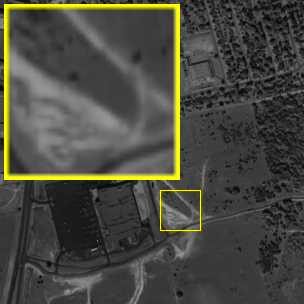}
		\\
		Noisy& TSLRLN&NGMeet& HSID-CNN&QRNN3D &T3SC& SST (Ours)
		\\ 
	\end{tabular}
	\vspace{-2mm}
	\caption{Visual comparison on real dataset Urban of band 105.}
	\vspace{-2mm}
	\label{fig:real-noisy} 
\end{figure*}

\noindent\textbf{Feature Representation.} 
In Figure~\ref{fig:feature}, we provide grayscale features maps after NLSA layer and GSA layer at different stages of our proposed Transformer, respectively. The output of GSA layer obtain more detailed information and structural texture than the outputs of NLSA layer. Since there is a close relationship between bands, different bands could complement each other by global spectral self-attention, resulting in a better-refined feature expression.

\noindent\textbf{Time Complexity.}  Parameters and GFLOPs are provided in
Table \ref{tab:flops} under $\sigma$$=$$50$. QRNN3D-L stands for QRNN3D with deeper layers and more channels compared to the one in \cite{wei20203}. Moreover, we also include Restormer ~\cite{zamir2022restormer} for comparison. Similarly, Restormer-L stands for a larger one. Besides, since HSID-CNN conducts denoising process in band-by-band manner, the GFLOPs is calculated by multiplying the band number and time that is required for one band. Our method achieves better result under comparable computation cost.

\subsection{Real Data Experiments}
\noindent\textbf{Setup.}
We evaluate our method on one real-world noisy HSI dataset named Urban. Urban dataset consists of 307$\times$307 pixels with 210 bands. 
Following \cite{bodrito2021trainable}, our Transformer and compared deep models are pre-trained on the APEX~\cite{itten2008apex} dataset, which has similar spectral coverage and band number to Urban dataset.

\begin{table}[t]
	\centering
	
	\setlength{\tabcolsep}{4.2mm}
	\renewcommand\arraystretch{0.7}
	
	\resizebox{\columnwidth}{!}{
		\begin{tabular}{c|c|c|c}
			\bottomrule
			Method & Param (M) & GFLOPs & PSNR (dB)\bigstrut\\
			\hline
			\hline
			QRNN3D&    0.86   &   19.6      & 39.70 \bigstrut\\
			\hline
			QRNN3D-L &    1.34   &    30.6      & 39.82 \bigstrut\\
			\hline
			HSID-CNN &  0.40     &  50.8        &  37.72\bigstrut\\
			\hline
			T3SC & 0.83      &    N/A       &  40.39\bigstrut\\
			\hline
			Restormer&    26.15   &  9.5      & 40.53 \bigstrut\\
			\hline
			Restormer-L&    45.14    &  21.9         &40.68  \bigstrut\\
			\hline
			SST (Ours)  &   4.14    &   20.7    & 41.09 \bigstrut\\
			\toprule
		\end{tabular}%
	}
	\vspace{-2mm}
	\caption{Model complexity comparisons.}
	\label{tab:flops}%
\end{table}%
\noindent\textbf{Visual Comparison.}
Since there is no clean image for real data, 
we only present grayscale images before denoising and after denoising to visually evaluate competing methods. The visual comparison results of one noisy band on Urban are shown in Figure \ref{fig:real-noisy}.
We can observe that the original HSI suffers from complex noise, which seriously affects the image quality. NGMeet seems to obtain over smoothed result and fails in preserving structural content and detail texture. Though QRNN3D and T3SC can recover a relatively complete image from the noisy band, the vertical stripes still exist in the restored images. Our method achieves satisfying denoising results and restores the image texture well.  

\begin{table}[t]
	\small
	\centering
	
	\renewcommand\tabcolsep{2.2mm}
	\renewcommand\arraystretch{0.68}
	\resizebox{\columnwidth}{!}{
		\begin{tabular}{c|c|c|c}
			\bottomrule
			Method &Params (M)& GFLOPs& PSNR (dB) \bigstrut\\
			\hline
			\hline
			w/o NLSA&    3.00   &    14.3   &   34.67      \bigstrut\\
			\hline
			w/o GSA&    2.98     &   13.1    &   40.44    \bigstrut\\
			\hline
			NLSA-NLSA  &     4.23 &   20.1      &  40.56 \bigstrut\\
			\hline
			GSA-GSA &   4.08    &   21.4    &   39.82      \bigstrut\\
			\hline
			GSA-NLSA  & 4.14 & 20.7 &40.69 \bigstrut\\
			\hline
			NLSA-GSA (Ours) &   4.14     &  20.7          & 41.09 \bigstrut\\
			\toprule
		\end{tabular}%
	}
	\vspace{-2mm}
	\caption{Ablation study related to the effectiveness of our proposed spatial-spectral multi-head self-attention.}
	\label{tab:ablation}%
\end{table}%

\begin{table}[htbp]
	\small
	\centering
	\renewcommand\tabcolsep{3.5mm}
	\renewcommand\arraystretch{0.75}
	\resizebox{\columnwidth}{!}{
		\begin{tabular}{c|c|c|c|c}
			\toprule
			Window Size & 2     & 4     & 8     & 16 \bigstrut\\
			\hline
			\hline
			GFLOPs & 16.87 & 17.50 & 20.70 & 30.24 \bigstrut\\
			\hline
			PSNR (dB) & 42.13 & 42.38 & 42.57 & 42.59 \bigstrut\\
			\hline
			SSIM & 0.9951 & 0.9953 & 0.9955 & 0.9955 \bigstrut\\
			\bottomrule
		\end{tabular}%
	}
	\vspace{-2mm}
	\caption{Analysis on the effect of window size.}
	\label{tab:window_size}%
\end{table}%

\subsection{Ablation Study}
To verify the effectiveness of our method, we perform ablation studies with noise level $\sigma$$=$$50$ on ICVL dataset. 

\noindent\textbf{Component Analysis.} In Table \ref{tab:ablation}, we investigate the effect of subcomponents in SSMA module, which includes NLSA layer and GSA layer. Without NLSA or GSA, the PSNR is 0.5 dB lower. 
To exclude the influence of computation complexity, we replace GSA layer with NLSA layer, resulting in a NLSA-NLSA module. Similarly, we also perform the experiment on the GSA-GSA module. It can be seen that under close computation cost, our proposed module still achieves the best performance, which suggests the higher efficiency of our proposed attention module. GSA-GSA obtained worst result among the last four methods. With NLSA layer before GSA layer, the feature extraction of GSA is more reliable. 

\noindent\textbf{Hyperparamter Analysis.}
To investigate the influence of window size $M$ in NLSA layer, we conduct experiments under different size of $M$ in Table~\ref{tab:window_size}. As $M$ increases, the network gets higher performance with larger computation cost. To make a better trade-off between performance and computation cost, we choose $M$=8 in our experiments.
\section{Conclusion}
In this paper, we propose a Spatial-Spectral Transformer for hyperspectral image denoising. 
The proposed Transformer considers both spatial similarity and spectral correlation in HSIs through the
spatial non-local self-attention and spectral global self-attention. 
The spatial non-local self-attention exploits the coarse features beyond neighboring pixels. Then, 
the spectral self-attention enriches the representation with more details. Extensive experiments verify the superiority of our method over state-of-the-art methods under various noise degradations quantitatively and visually. In the future, it is worth investigating how to use noise estimation as guidance to boost the denoising performance of Transformer.

\balance
\bibliography{aaai23}

\end{document}